\documentclass[runningheads]{llncs}
\usepackage{subcaption}
\usepackage{graphicx}
\usepackage{hyperref}
\usepackage{todonotes}
\usepackage{amsmath}
\usepackage{amssymb}
\usepackage{siunitx}

\begin{document}
\title{Prediction of fish location by combining fisheries data and sea bottom temperature forecasting\thanks{This work has received funding from the European Union’s Horizon 2020 research and innovation program under grant agreement NO.825355.}}
\titlerunning{Prediction of fish location by combining fisheries data and temperature}
%
\author{Matthieu Ospici\inst{1}\orcidID{0000-0002-7816-721X} \and Klaas Sys\inst{2}\orcidID{0000-0001-8356-460X} \and
Sophie~Guegan-Marat\inst{1}\orcidID{0000-0002-7187-4288}}
\authorrunning{M. Ospici et al.}
%
\institute{Atos, CVLab, 38130 Echirolles, France\\
\email{\{matthieu.ospici,sophie.gueganmarat\}@atos.net}
\and
Flanders Research Institute for Agriculture, Fisheries and Food, Animal Sciences Unit, Ankerstraat 1, 8400 Oostende, Belgium\\
\email{klaas.sys@ilvo.vlaanderen.be}}

\maketitle              
\begin{abstract}
This paper combines fisheries dependent data and environmental data to be used in a machine learning pipeline to predict the spatio-temporal abundance of two species (plaice and sole) commonly caught by the Belgian fishery in the North Sea. By combining fisheries related features with environmental data, sea bottom temperature derived from remote sensing, a higher accuracy can be achieved. In a forecast setting, the predictive accuracy is further improved by predicting, using a recurrent deep neural network, the sea bottom temperature up to four days in advance instead of relying on the last previous temperature measurement.

\end{abstract}

\section{Introduction}
Spatio-temporal modeling of catch or landings per unit effort data has a long history in fisheries sciences. Such models are mainly used to standardize fisheries dependent (i.e. landing and effort data from commercial fisheries) or fisheries independent (catch data by haul from scientific research cruises) information in order to derive an index of abundance~\cite{maunder_2004}. Such an index is typically used to inform population dynamic models used to assess the exploitation status of fish stocks. 

Yet few applications exist in which spatio-temporal models, fitted to catch or landings and effort data, are used to forecast the spatio-temporal dynamics of fish species. Such applications could be very valuable in the context of dynamic ocean management, a new fisheries management paradigm defined as ‘management that changes rapidly in space and time in response to the shifting nature of the ocean and its users based on the integration of new biological, oceanographic, social and/or economic data in near real-time’~\cite{maxwell_2015}. A notable example of dynamic ocean management are Real Time Incentives (RTIs), a system that allows to reduce by-catches of fish by providing weekly maps with tariffs set according the expected catches of the fishery for a given set of species~\cite{kraak_2015}. Clearly, applications of dynamic ocean management typically build on the fusion of alternative data sources, i.e. remote sensing, and advanced analytical processing and modeling techniques.

In this paper, we combine common fisheries dependent data sources, being daily landings reported by fishers through electronic logbooks and data on vessel activity provided through the Vessel Monitoring System (VMS), and environmental data (sea bottom temperature). With these data, we use a computer vision and machine learning pipeline to predict the spatio-temporal abundance of two commercially important species of the Belgian fishery, sole (\textit{Solea solea}) and plaice (\textit{Pleuronectes platessa}).

\section{Related works}
\emph{fish distribution forecasting} Different applications exist that forecast the spatio-temporal distribution of fish in relation to environmental conditions. Many of these application build on suitable habitat models in which so called environmental envelopes are constructed from species occurrence data or expert knowledge. These envelopes, constrained to some predefined shape (i.e. trapezoidal~\cite{kaschner_2006}, shape constrained GAMS~\cite{citores_2020}), show the probability of species occurrence in response to a set of environmental variables and can as such be used to predict suitable habitats. Typically, the environmental envelopes are rather static and are used to make long term prediction of changes in species distribution for a given climate scenario.

An alternative approach was developed for the EcoCast application in which machine learning models are used to predict the daily probability of occurrence of swordfish, sea lions, leatherback turtles and blue sharks off the coast of Oregon and California~\cite{hazen_2018}. These individual species maps are combined to identify suitable areas for fishing. The approach applies boosted regression trees on historical observer (1991 - 2014) (presence/absence) and tracking (2001 - 2009) data derived from tagging experiments combined with multiple environmental data to model the probability of species occurrence. This model uses the most recent observations (often the previous day) of the environmental parameters used in the model to make predictions for the current day. As such, it is assumed that the most recent environmental conditions are similar to the current environmental conditions, a shortcoming we aim to address in this study.

\emph{temperature forecasting with machine learning}  Many approaches have been specifically proposed to predict temperature from satellite data with recurrent deep learning methods~\cite{xiao_spatiotemporal_2019,kim_sea_2020,qiao_sea_2021,nipslf}. Simultaneously, related approaches, designed to predict images from videos, have been proposed~\cite{wang_predrnn_nodate,fpred1}. To our knowledge, none of these works have been applied to the prediction of water temperature at the sea bottom.

\section{Proposed framework}
To predict the probability of fish presence at a given location at sea, we propose a framework that consists of two main building blocks. The first is a deep learning model, trained on satellite images, made to predict the sea bottom temperature. The second block is a machine learning (gradient boosting) model, which uses the predicted temperature as input plus a set of features from a dataset that comprises information on landings of particular fish species at particular locations.

\subsection{Datasets used}
\subsubsection{Satellite data}
The satellite data of sea bottom temperature used in this work has been collected using the Copernicus Marine Environment Monitoring Service (CMEMS). Sea bottom temperatures are not directly observed but generated by the NEMO (Nucleus for European Modelling of the Ocean) ocean model, using the 3DVar NEMOVAR system to assimilate observations.  

\begin{figure}
\centering
\includegraphics[width=0.7\textwidth]{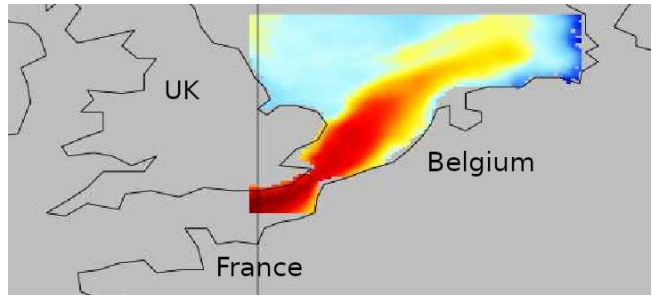}
\caption{The region of interest is an area located between north of France and south of UK (the area is in color in the picture)}\label{fig:aera}
\centering
\end{figure}

The region of interest for our experiments is in a rectangular area with latitude ranging from 50.07 to 55.33 and longitude ranging from 0.22 to 8.56. The resolution of the latitude and longitude is 80x80. Figure~\ref{fig:aera} shows the location of this area on the earth. We collected data from the years 2006 to 2020, giving a total of 5295 successive days of bottom temperature.

\subsubsection{Fisheries data}

A dataset that contains species landings of the Belgian fishing fleet was compiled from electronic logbook and VMS data. The electronic logbook data comprise daily information on fish landings, a description of the fishing activity (gear and mesh size used, ICES statistical rectangle, fishing trip identifier, and day), and information on the fishing vessel (vessel identifier, vessel tonnage, engine power and length). The landing and effort data were combined according an agreed procedure as described in~\cite{hintzen}. By merging both datasets, a final dataset with landing data at a higher spatio-temporal resolution is generated.
Data was available from 2006 up to 2020. In total the dataset comprised \num{1684560} observations.

\subsection{Predicting bottom temperature with deep learning}
The first component of our framework is a deep learning model capable of forecasting the bottom temperature from a temperature history.

\subsubsection{Problem formulation}
The bottom temperature for a given timestamp can be represented by a matrix $BT$ of dimensions $M \times N$, where $M$ denotes the resolution in longitude space and $N$ the resolution in latitude space. Therefore, $BT[i][j]$ is the numerical value of the bottom temperature at coordinate $i,j$ in degrees Celsius.

We can then construct a sequence of consecutive bottom temperatures for a given time period. Let $BT_t$ be the temperature matrix at time $t$. The sequence is then the following:
\begin{equation}
  BT_{0},BT_{1}, \cdots, BT_{h-1},    BT_{h}, \cdots , BT_{h+p-1}
  \label{eq:seq}
\end{equation}

The length of the sequence is $l = p + h$ with $p$ the number of matrices to predict and $h$ the number of available histories. Our goal is therefore to predict $p$ matrices of consecutive bottom temperatures given a history of $h$ measures.


\subsubsection{Creation of the sequences and processing of land areas}


The raw dataset contains a large sequence of size $nb_{R}=5295$ bottom temperature matrices. The sequences needed by the neural network are built by advancing step by step in the raw dataset. Thus, the dataset size used for training, which is the number of sequences of length $l$ can be calculated as follows: $n_{sequence} = nb_{R}-l$


\label{txt:minus5}
Then, the data are normalized by subtracting each value by the mean value of the dataset and dividing by its standard deviation. Our region of interest contains land, which corresponds to NaN values in the dataset. These NaN values are replaced, after normalization, by a high negative value ($-5$ in our implementation).

\subsubsection{Recurrent neural network}
Our work is based on the framework PredRNN++ \cite{wang_predrnn_nodate}, we build our forecasting model on the building blocs introduced in~\cite{wang_predrnn_nodate} which are the recurrent units called Causal LSTM and the Gradient Highway Unit.
We motivate our choice because this framework can handle short-term dependencies very efficiently and outperforms traditional approaches, often used in satellite data forecasting, such as LSTM or Convolutional LSTM~\cite{xiao_spatiotemporal_2019,kim_sea_2020,moskolai_application_2021}.

\subsection{Fish prediction with gradient boosting}
The fisheries dataset, which contains a substantial history of fishing operations in the North Sea, is used to build a machine learning model that aims to predict the probability of fish presence in the sea.

Since plaice and sole are bottom dwelling species that have a narrow thermal preference range, the temperature of the sea floor is assumed to influence the location of fish. Hence, we use the temperature prediction model presented in the previous section among the features used by our machine learning model. Therefore, the resolution of the fish prediction is the same as the resolution of the temperature map. As discussed in the previous section, the resolution of the temperature map is 80x80, so the resolution of the fish prediction is the same.

\subsubsection{Gradient boosting to predict potential fishing zone}
Since the fisheries dataset contains many different types of features, we chose a decision tree-based machine learning algorithm. Specifically, we used the lightGBM~\cite{ke_lightgbm_2017} framework with GPU acceleration~\cite{zhang_gpu-acceleration_nodate}. We build two models for two species of fish: one to predict the probability of presence of plaice, and one to predict the probability of presence of sole. The output of the models is then a probability of fish presence for each point of the 80x80 map.

\subsubsection{Features selection}
From the original dataset, we build seven features to train the machine learning models: longitude ($i$), latitude ($ii$),  bottom temperature ($iii$) and two features for the day ($iv$, $v$) and two for the year ($vi$, $vii$).  Indeed, we transform day and month, which are cyclic in nature, into two new features using a sine and cosine transformation~\cite{cyclic}. Furthermore, because the fish are predicted in an 80x80 resolution grid, the latitude and longitude correspond to the coordinates of the point in which the prediction is made and are therefore integers ranging from 0 to 79. 



\section{Experimental settings}

\subsection{Train-test-evaluation splits}
The recurrent neural network is trained and evaluated on satellite data from sequences defined in section \ref{txt:minus5}. We follow the evaluation procedure described in \cite{xiao_spatiotemporal_2019}, the training, validation, and test sets are chosen to follow each other in chronological order. In our study, the training set consists of the years from 2006 to 2018, and the evaluation set and test set are formed from the years 2019 to 2020. 
The plaice and sole gradient boosting models are trained on both satellite and fisheries dataset with a similar train-test-evaluation split. We take the years from 2019 to 2020 as test and evaluation sets.

\subsection{Hyperparameters}
The hyperparameters of the recurrent neural network for temperature forecasting are as follows. The input and output lengths are set to 4, which means that we predict 4 days with a history of 4 days. The optimization is performed with Adam~\cite{DBLP:journals/corr/KingmaB14} with a learning rate of 0.001. The loss function used in our experiments is the $L1+L2$ function which gives the best performance compared to using $L1$ or $L2$ functions only. The batch size is set to 8. Following the neural network architecture introduced in \cite{wang_predrnn_nodate}, each layer is built with 4 causal LSTMs with 128, 64, 64, 64 channels and a 128 channel gradient highway unit. In all recurrent units, the convolution filter size is set to 5.

The two lightGBM models use the standard hyperparameters of the framework~\cite{ke_lightgbm_2017} with a number of leaves set to 23. The objective is binary classification and the loss used is the log-loss. 

\subsection{Metrics}
We evaluate the temperature forecasting model using two metrics well suited to regression problems: MAE (mean absolute error) and  RMSE (root-mean-square error).
For the fish prediction models, we use the F1 score, which is well suited to binary classification problems.
\begin{equation*}
  \mathit{RMSE}=\sqrt{\frac{1}{mn} \sum_{i=1}^m \sum_{j=1}^n (x_{ij}-y_{ij})^2}
  \quad\text{,}\quad 
  \mathit{MAE}=\frac{1}{mn} \sum_{i=1}^m \sum_{j=1}^n |x_{ij}-y_{ij}|
\end{equation*}
\begin{equation*}
  \mathit{F1} = \frac{2*Precision*Recall}{Precision+Recall}
\end{equation*}
$m$ and $n$ correspond to the latitude and longitude resolutions ($m=n=80$). Since we have land in our study area, we make sure that for both RMSE and MAE, the coordinates $i,j$ correspond only to points that are in the sea.

\section{Performance evaluation}
To measure the performance of our framework, we first evaluate the performance of its two component and then the performance of the entire framework. For the fish prediction models, we also verify that they have captured the link, which exists in the marine environment, between bottom temperature and fish abundance.

\subsection{Temperature forecasting performance}
We first ensure that the forecasting performance of our model is better than simply taking the last known day as a predictor of future days' temperature.
Then, because we have chosen to set the temperature value of the areas with land to a high negative value, we validate this approach.

\subsubsection{Last know day estimator compared to the proposed forecasting model}
\paragraph{Motivation}

For a given sequence such as the one in equation~(\ref{eq:seq}), we call the last day estimator the temperature matrix $BT_{h-1}$. We can then calculate the error (RMSE and MAE) between this matrix and the matrices $BT_{h}\cdots BT_{h+p-1}$.

 Similarly to \cite{nipslf}, we perform this experiment because we observed that the variation of the temperature is extremely low between the different days. For this reason, it would be easy to develop a forecasting algorithm with a low error rate by taking only the last know day as prediction. It is then essential to ensure that our forecast model is able to outperform the last known day estimator.

\begin{figure}
\centering
\begin{subfigure}{0.49\textwidth}
    \includegraphics[width=\textwidth]{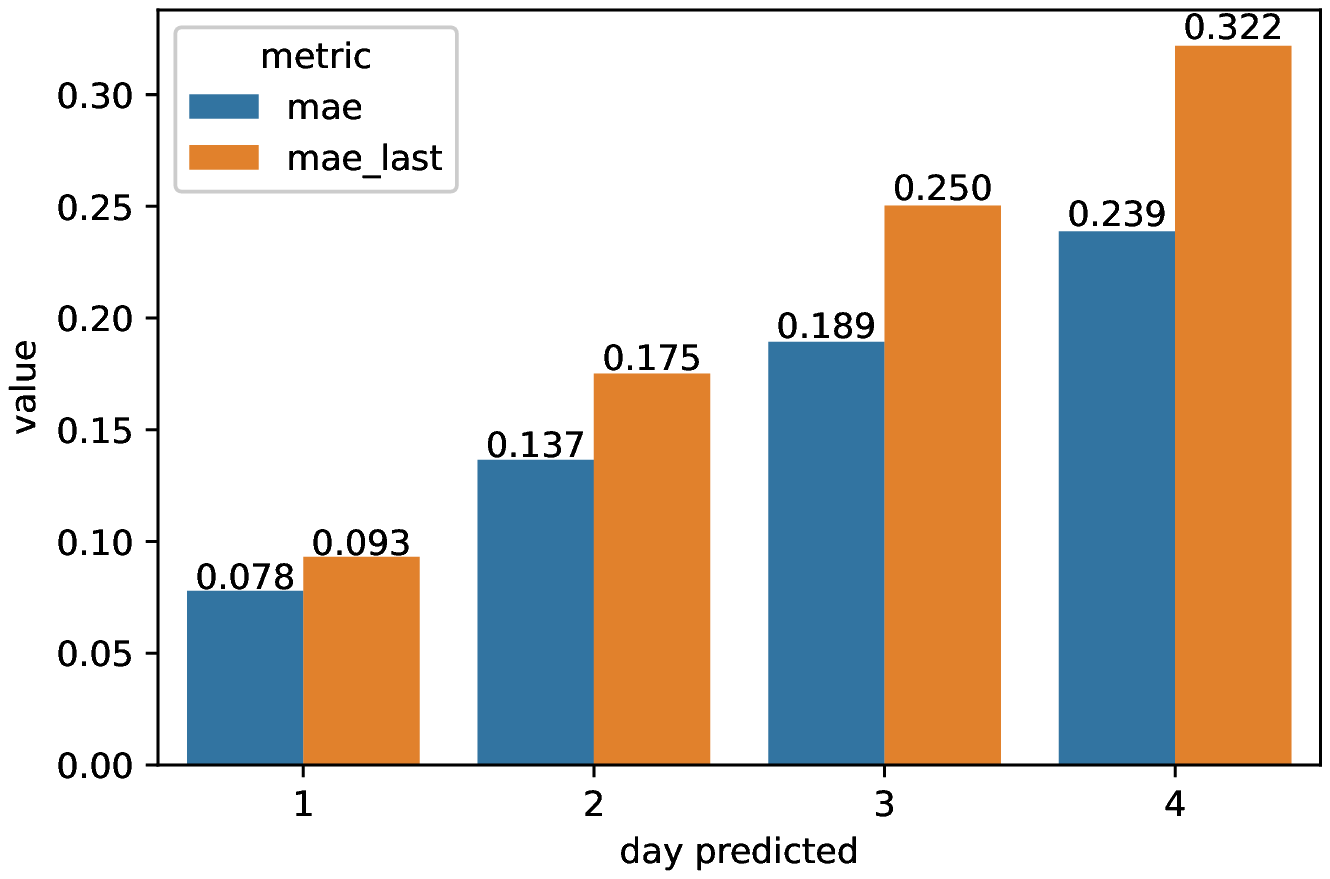}
    \caption{MAE}
    \label{fig:mae}
\end{subfigure}
\hfill
\begin{subfigure}{0.49\textwidth}
    \includegraphics[width=\textwidth]{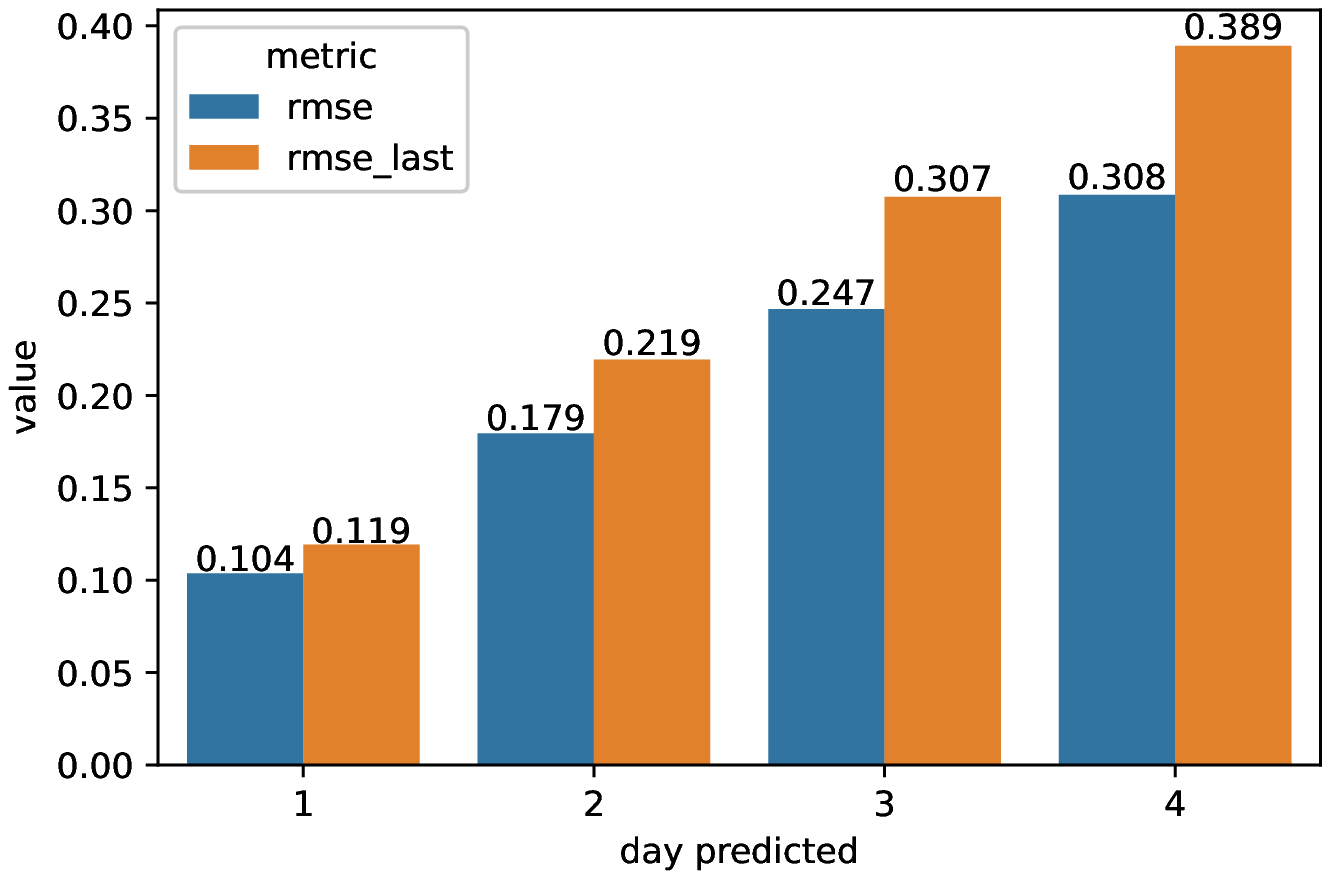}
    \caption{RMSE}
    \label{fig:rmse}
\end{subfigure}
\caption{For both figures, the error is calculated between the forecast and the ground truth (blue bars) and between the last day estimator and the ground truth (orange). The lowest values are the best.}
\label{fig:mae_rmse}
\end{figure}

\paragraph{Results}
The errors obtained by simply using the last known day are shown in orange in Figure~\ref{fig:mae} for MAE and Figure~\ref{fig:rmse} for RMSE. 
As expected, using the last known day as the temperature estimator for the following days results in a relatively small error ($0.10^\circ C$ of MAE) for the first day. The error increases significantly for the following days (for predicting the fourth day, the error is $0.3^\circ C$).

In both figures \ref{fig:mae} and \ref{fig:rmse}, we plot the error for the day predicted by our model (in blue). 
We can observe here that our model outperforms the estimator of the last known day with both RMSE and MAE metrics for all days. Furthermore, the percentage of error between the estimator of the last known day and the value predicted by our model increases when we progress in the predictions. The error (in terms of RMSE) is 14.4\% higher for the estimator of the last known day for day 1 and 26.3\% for day 4.

\subsubsection{Repartition of the errors}
\begin{figure}[h]
\centering
\begin{subfigure}{0.24\textwidth}
    \includegraphics[width=\textwidth]{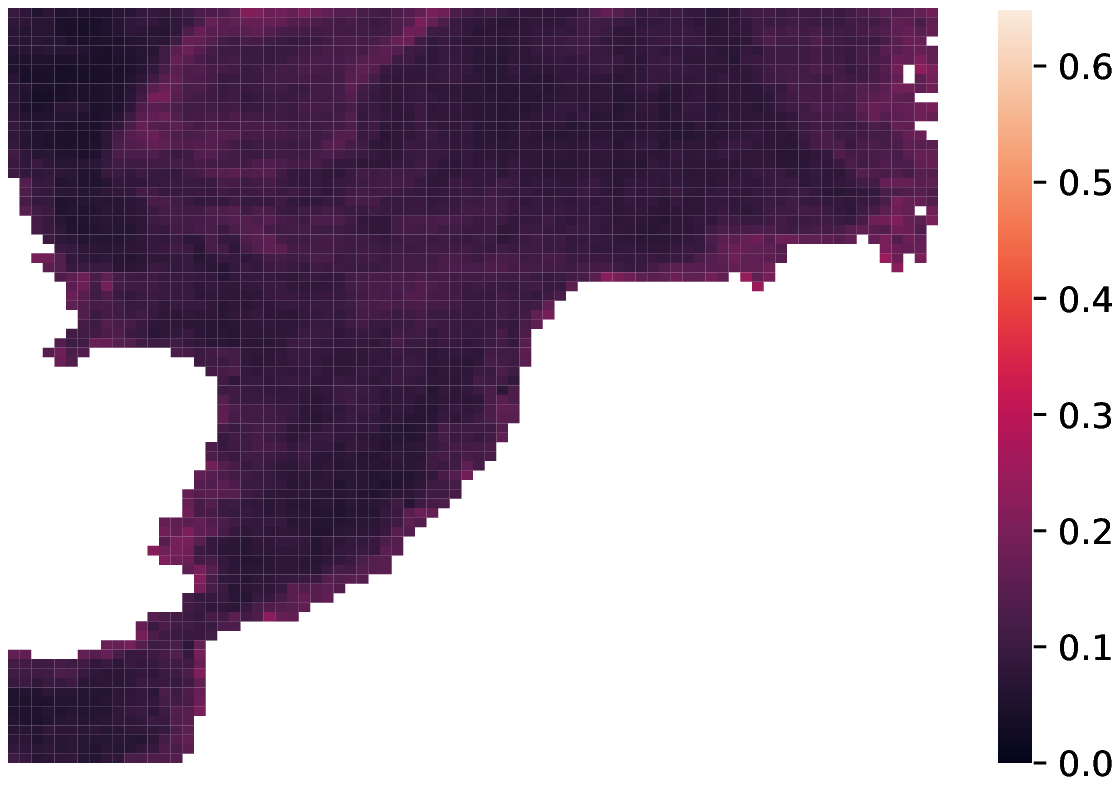}
    \caption{day 1}
    \label{fig:pred_1}
\end{subfigure}
\hfill
\begin{subfigure}{0.24\textwidth}
    \includegraphics[width=\textwidth]{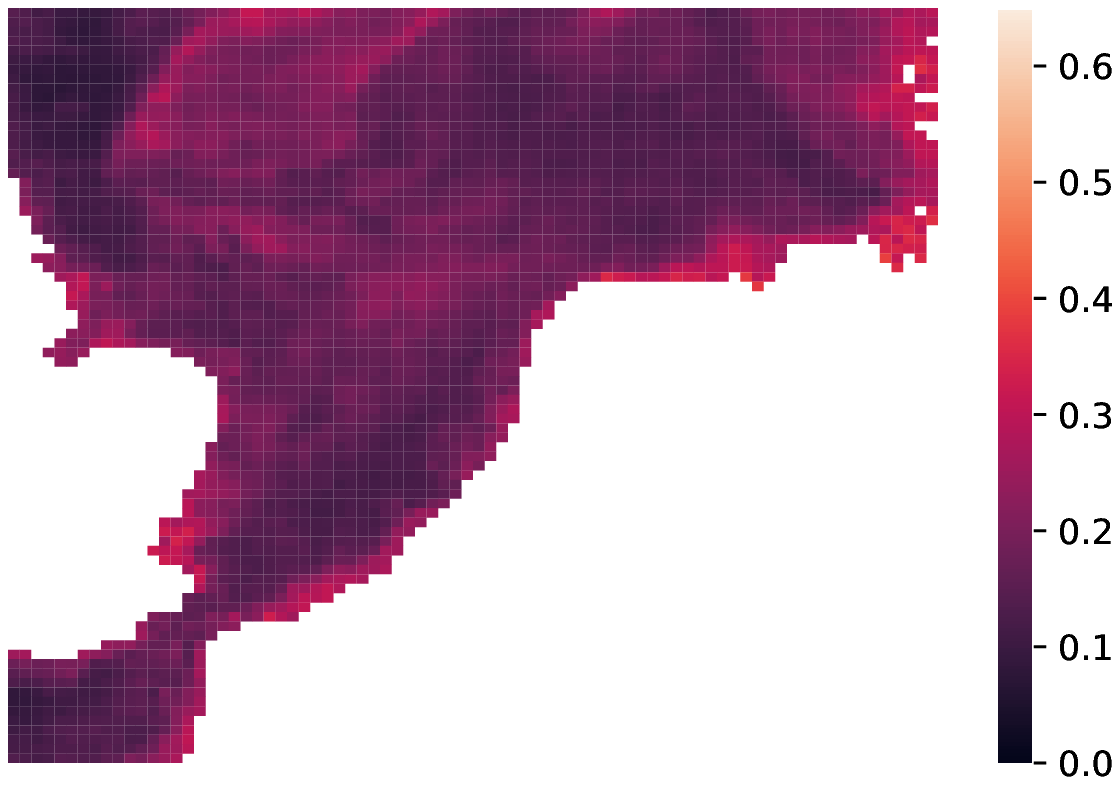}
    \caption{day 2}
    \label{fig:pred_2}
\end{subfigure}
\hfill
\begin{subfigure}{0.24\textwidth}
    \includegraphics[width=\textwidth]{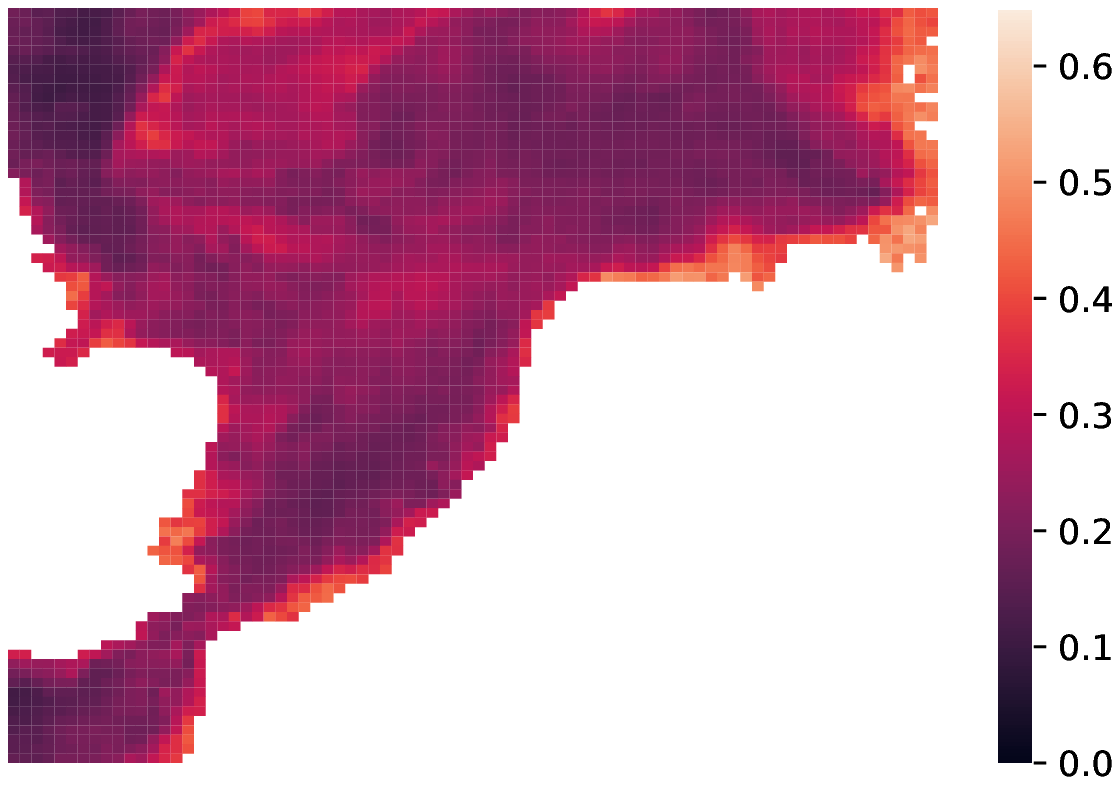}
    \caption{day 3}
    \label{fig:pred_3}
\end{subfigure}
\hfill
\begin{subfigure}{0.24\textwidth}
    \includegraphics[width=\textwidth]{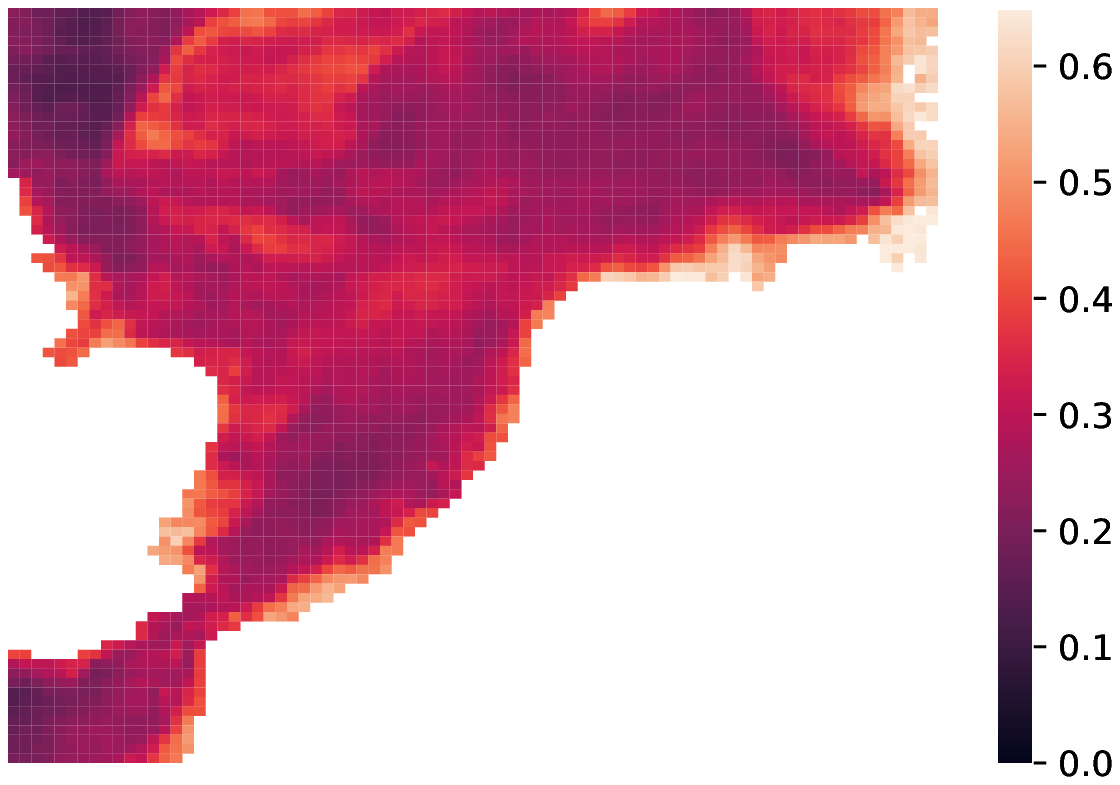}
    \caption{day 4}
    \label{fig:pred_4}
\end{subfigure}  
\caption{RMSE between ground truth and prediction for day 1 to 4. RMSE is computed for each points (80x80) in the area of interest. The white area is the land.}
\label{fig:pred}
\end{figure}

During the preprocessing of the dataset, we replace the NaN values (which represent the land) by a high negative value. To validate this approach we compute the RMSE between the predicted temperatures and the ground truth for days 1-4 and we show the RMSE error for each cell of the error matrix. Results are depicted in Figures~\ref{fig:pred_1} to \ref{fig:pred_4}. 

For the predicted days, the errors are fairly well distributed in all areas. The errors around the interface between the sea and the land are comparable to those of the other zones, which validates our approach.

It should be noted, however, that these areas at the interface are among the areas that concentrate the most errors, especially when we advance in the prediction. This can be explained by the fact that these are shallow areas where the temperature variations are the most important from one day to another. We illustrate this on the figure \ref{fig:last-estim}. We calculate the location of the RMSE errors between the Last know day estimator and the ground truth of day 4. There is no prediction here, we only compare the ground-truth. The largest errors correspond to the largest temperature variations. This confirm that the areas where the prediction errors are the largest correspond to the areas where the temperature variations are the most important.

\subsection{Influence of bottom temperature for fish prediction}
\subsubsection{Link between temperature and quality of prediction}
We train both fish prediction models with the fisheries dataset. To validate the relevance of using bottom temperature for the prediction of sole and plaice presence, we perform two experiments. In the first, we train and test a model with and without the "bottom temperature" feature. The results are depicted in Table~\ref{tab1} under the columns ``w/ temp.'' and ``w/o temp.''. For this experiment, the bottom temperature is not predicted, we use the ground truth value from the Copernicus dataset.

According to the result presented Table~\ref{tab1}, using bottom temperature for both plaice and sole models increases the quality of the prediction. This justify our approach of using bottom temperature for fish presence prediction. 

\subsubsection{Link between temperature and abundance of fish}
\begin{figure}
\centering
\begin{minipage}{0.5\textwidth}
  \includegraphics[width=\linewidth]{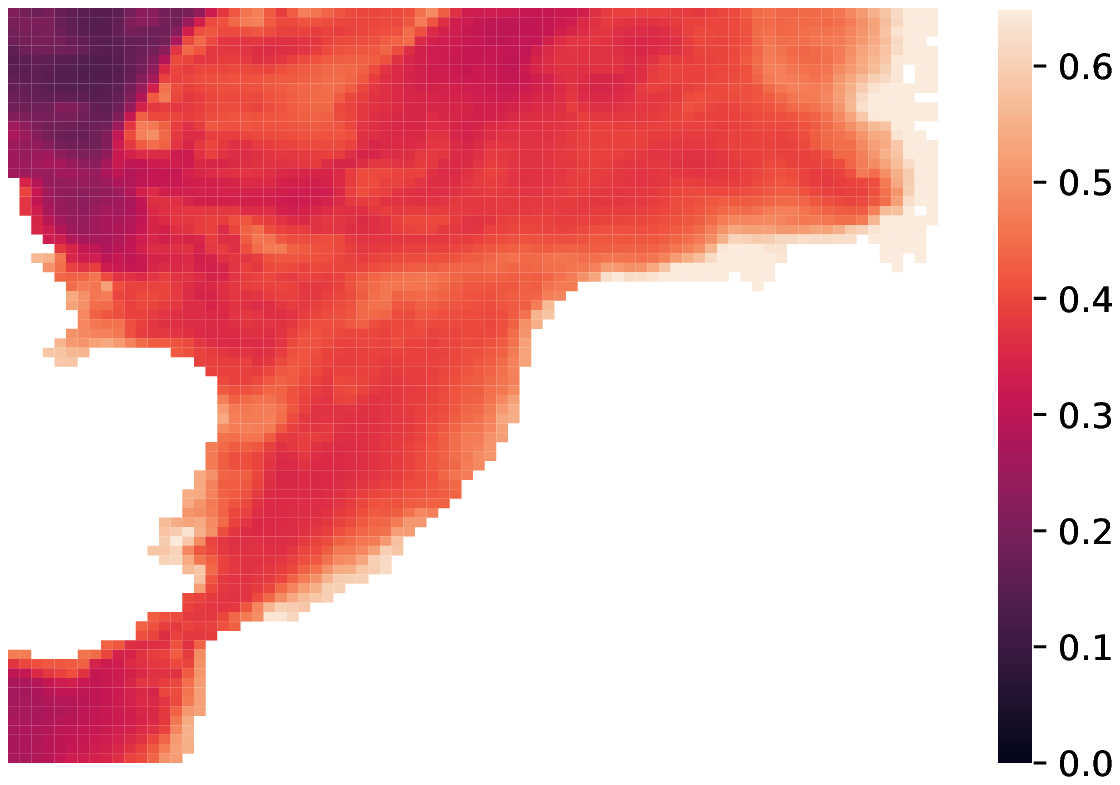}
  \captionsetup{width=0.8\linewidth}
  \captionof{figure}{RMSE errors between the Last know day estimator and the ground truth of day 4.}
  \label{fig:last-estim}
\end{minipage}%
\begin{minipage}{0.5\textwidth}
  \includegraphics[width=\textwidth]{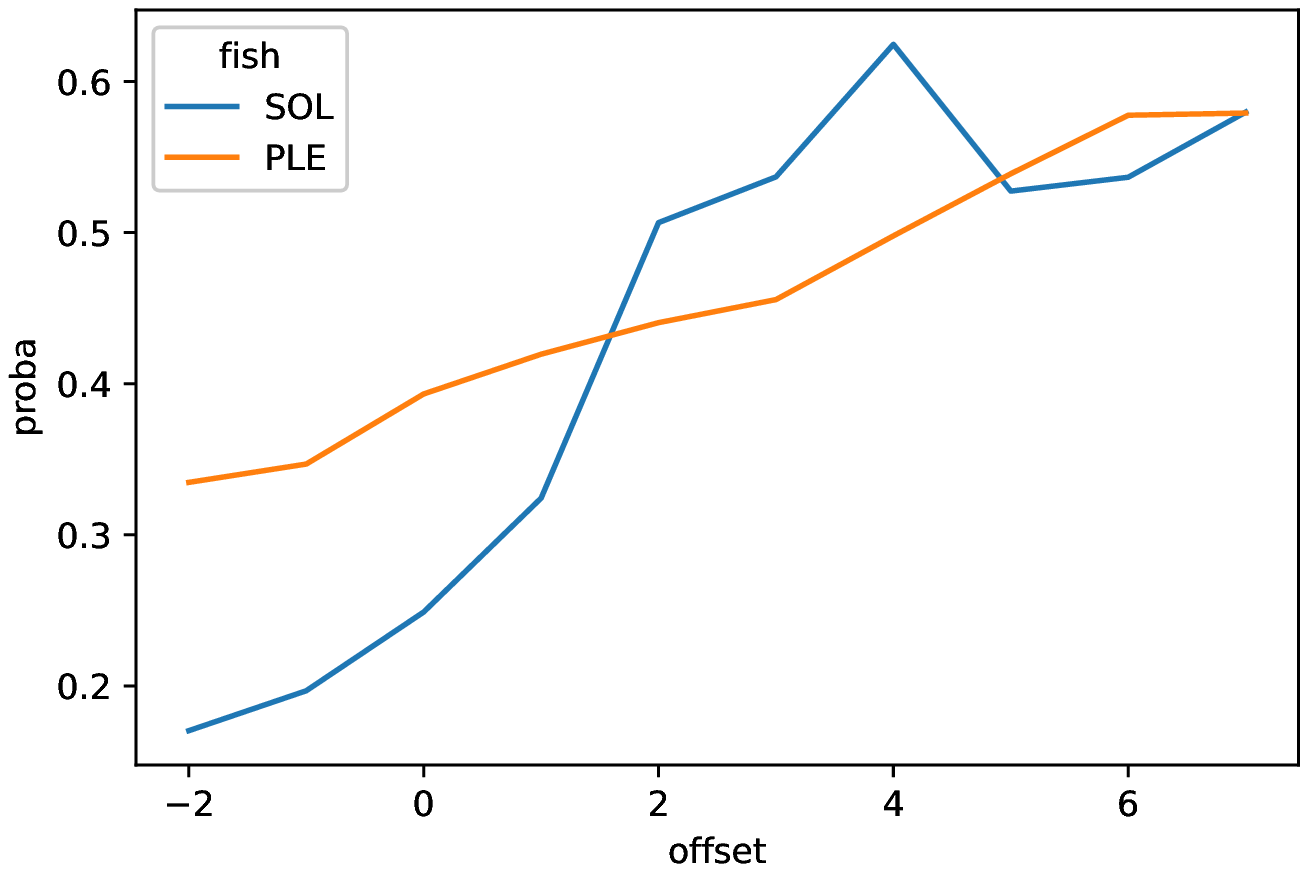}
  \captionsetup{width=0.8\linewidth}
  \captionof{figure}{Average probability of fish presence when an offset is added to bottom temperature.}
  \label{fig:tempincr}
\end{minipage}
\end{figure}

To more precisely measure how the model has captured the link between temperature and presence of fish, the following experiment was conducted. We added an offset, from $-2^\circ C$ to $7^\circ C$, to the bottom temperature of each grid point on a given day and computed the mean of probability of fish presence in the study area using the sole and plaice model. Results are depicted in Figure~\ref{fig:tempincr}. As shown, when we decrease (resp. increase) the temperature, the model predicts a lower (resp. higher) occurrence probability of fish on average. The stronger positive correlation between species occurrence and seawater temperature for sole compared to plaice can be explained by the fact that sole is a  Lusitanian species that prefers warmer water. Moreover, the study area comprises the northernmost habitat of the sole. In contrast, plaice is widely distributed over the entire North Sea which may explain why the response of species occurrence to temperature is less pronounced. 

\begin{table}
\caption{F1 score for plaice (PLE) and sole (SOL) models}
\label{tab1}
\center
\begin{tabular}{|l|c|c|c|c|c|c|}
\hline
Fish type &  w/  temp. & w/o  temp. & pred. day 1 & pred. day 2 & pred. day 3 & pred. day 4\\
\hline
SOL & 82.8 & 81.4 & 82.7& 82.7& 82.6& 82.4 \\
PLE & 82.0 & 80.6 & 81.8& 81.8& 81.8& 81.8\\
\hline
\end{tabular}
\end{table}

\subsection{Performance of the complete framework}
The accuracy of the fish prediction is evaluated using the predicted bottom temperature as follows: for each day $d$ of the fisheries data test set, we construct four predictions. For the prediction at day 1 (pred. day 1 in Table~\ref{tab1}), we construct a sequence $BT_{d-4},BT_{d-3}, \cdots, BT_{d-1}$ from the ground truth measurements. Thus, with our bottom temperature prediction model, we can predict day $d$ by taking the first day of the predicted sequence. For the prediction of day 2, we build a sequence $BT_{d-5},BT_{d-4}, \cdots, BT_{d-2}$ and take the second predicted day. We proceed in the same way for day 3 and day 4. We then measure the quality of the prediction in terms of F1 score when we use these different predictions as the temperature for day $d$. The results are presented in Table~\ref{tab1}.

Our experiments show that, even the prediction of temperature contain small errors in term of MAE or RMSE, using this predicted value shows excellent results for 1,2,3 and 4 days ahead. With a 4-day prediction we have performance extremely close to using a ground truth value for both sole and plaice.

\section{Discussion and Conclusion}
Our study showed how a pipeline of advanced analytical tools and data fusion can be used to make a short term forecast of species occurrence in a highly variable environment such as the marine ecosystem. 

We shown that a recurrent neural network building blocks designed to predict successive frames of a video can be efficient for sea bottom temperature forecasting. During the evaluation, we noticed that the variation in temperature is small and even using only the last known day as a predictor, we can have small errors. We therefore carefully ensured that our model outperformed this simple estimator. Moreover, our choice to replace the value of the coordinates containing land by a high negative value has been validated.

By enriching a fisheries dependent dataset with features on environmental conditions such as sea bottom temperature, a higher predictive accuracy of sole and plaice occurrence could be achieved with a gradient boosting algorithm. This accuracy was further increased by using the predicted sea bottom temperature that was inferred from a recurrent deep learning model. Since information on environmental conditions is currently available in near real time through satellite based earth monitoring programs such as Copernicus, this may provide an opportunity for practical applications of dynamic ocean management. 

Although the accuracy of sole and plaice occurrence could be increased by adding an environmental feature to the fisheries dataset, the overall accuracy can still be improved. It should be noted however that the quality of the fisheries dependent data is limited. The landing data reported by fishers are estimated with a fault tolerance of 20\%. Moreover, misreporting is known to occur wherewith species landings from one area are reported in another area. Finally, the landing data are reported on a daily basis and distributed over the GPS positions of a fishing vessel that were recorded during that day according~\cite{hintzen}. To do so, an equal distribution of landings over all fishing position is assumed which is unlikely to be the case in reality. As more data will become available in the future, and alternative catch monitoring techniques, e.g. image analysis, will be implemented, the predictive accuracy of fish occurrence models relying on fisheries dependent information is likely to increase.


%
%
%
\bibliographystyle{splncs04}
\bibliography{cybele_pilot8}

\begin{thebibliography}{10}
\providecommand{\url}[1]{\texttt{#1}}
\providecommand{\urlprefix}{URL }
\providecommand{\doi}[1]{https://doi.org/#1}

\bibitem{cyclic}
Chakraborty, D., Elzarka, H.: Advanced machine learning techniques for building
  performance simulation: a comparative analysis. Journal of Building
  Performance Simulation  \textbf{12},  1--15 (07 2018)

\bibitem{citores_2020}
Citoresa, L., Ibaibarriaga, L., Lee, D.J., Brewer, M., Santos, M., Chust, G.:
  Modelling species presence–absence in the ecological niche theory framework
  using shape-constrained generalized additive models. Ecological Modelling
  \textbf{418} (2020)

\bibitem{hazen_2018}
Hazen, E.L., Scales, K.L., Maxwell, S.M., Briscoe, D.K., Welch, H., Bograd,
  S.J., Bailey, H., Benson, S., Eguchi, T., Dewar, H., Kohin, S., Costa, D.P.,
  Crowder, L.B., Lewison, R.L.: A dynamic ocean management tool to reduce
  bycatch and support sustainable fisheries. Science Advances  \textbf{4 -(5)},
  ~1--8 (May 2018)

\bibitem{hintzen}
Hintzen, N.T., Bastardie, F., Beare, D., Piet, G.J., Ulrich, C., Deporte, N.,
  Egekvist, J., Degel, H.: Vmstools: Open-source software for the processing,
  analysis and visualization of fisheries logbook and vms data. Fisheries
  Research  \textbf{115 - 116},  31--43 (Nov 2012)

\bibitem{kaschner_2006}
Kaschner, K., Watson, R., Trites, A.W., Pauly, D.: Mapping world-wide
  distributions of marine mammal species using a relative environmental
  suitability (res) model. Marine Ecology Progress Series  \textbf{316},
  285--310 (Jul 2006)

\bibitem{ke_lightgbm_2017}
Ke, G., Meng, Q., Finley, T., Wang, T., Chen, W., Ma, W., Ye, Q., Liu, T.Y.:
  {LightGBM}: {A} {Highly} {Efficient} {Gradient} {Boosting} {Decision} {Tree}.
  In: Guyon, I., Luxburg, U.V., Bengio, S., Wallach, H., Fergus, R.,
  Vishwanathan, S., Garnett, R. (eds.) Advances in {Neural} {Information}
  {Processing} {Systems}. vol.~30. Curran Associates, Inc. (2017)

\bibitem{kim_sea_2020}
Kim, M., Yang, H., Kim, J.: Sea {Surface} {Temperature} and {High} {Water}
  {Temperature} {Occurrence} {Prediction} {Using} a {Long} {Short}-{Term}
  {Memory} {Model}. Remote Sensing  \textbf{12}(21), ~3654 (Jan 2020), number:
  21 Publisher: Multidisciplinary Digital Publishing Institute

\bibitem{DBLP:journals/corr/KingmaB14}
Kingma, D.P., Ba, J.: Adam: {A} method for stochastic optimization. In: Bengio,
  Y., LeCun, Y. (eds.) 3rd International Conference on Learning
  Representations, {ICLR} 2015, San Diego, CA, USA, May 7-9, 2015, Conference
  Track Proceedings (2015)

\bibitem{kraak_2015}
Kraak, S.B., Reid, D.G., Bal, G., Barkai, A., Codling, E.A., Kelly, C.J.,
  Rogan, E.: Rti (“real-time incentives”) outperforms traditional
  management in a simulated mixed fishery and cases incorporating protection of
  vulnerable species and areas. Fisheries Research  \textbf{172},  209--224
  (Dec 2015)

\bibitem{maunder_2004}
Maunder, M.N., Punt, A.E.: Standardizing catch and effort data: a review of
  recent approaches. Fisheries Research  \textbf{70 (2-3)},  141--159 (Dec
  2004)

\bibitem{maxwell_2015}
Maxwell, S.M., Hazen, E.L., Lewison, R.L., Dunn, D.C., Bailey, H., Bograd,
  S.J., Briscoe, D.K., Fossette, S., Hobday, A.J., Bennett, M., Benson, S.,
  Caldwell, M.R., Costa, D.P., Dewar, H., Eguchi, T., Hazen, L., Kohin, S.,
  Sippel, T., Crowder, L.B.: Dynamic ocean management: Defining and
  conceptualizing real-time management of the ocean. Marine Policy
  \textbf{58},  42--50 (Aug 2015)

\bibitem{moskolai_application_2021}
Moskolaï, W., Abdou, W., Dipanda, A., Kolyang: Application of {Deep}
  {Learning} {Architectures} for {Satellite} {Image} {Time} {Series}
  {Prediction}: {A} {Review}. Remote Sensing  (2021)

\bibitem{qiao_sea_2021}
Qiao, B., Wu, Z., Tang, Z., Wu, G.: Sea {Surface} {Temperature} {Prediction}
  {Approach} {Based} on {3D} {CNN} and {LSTM} with {Attention} {Mechanism}. In:
  2021 23rd {International} {Conference} on {Advanced} {Communication}
  {Technology} ({ICACT}). pp. 342--347 (Feb 2021), iSSN: 1738-9445

\bibitem{nipslf}
Shi, X., Gao, Z., Lausen, L., Wang, H., Yeung, D.Y., Wong, W.k., Woo, W.c.:
  Deep learning for precipitation nowcasting: A benchmark and a new model. In:
  Proceedings of the 31st International Conference on Neural Information
  Processing Systems. p. 5622–5632. NIPS'17, Curran Associates Inc., Red
  Hook, NY, USA (2017)

\bibitem{fpred1}
Su, J., Byeon, W., Kossaifi, J., Huang, F., Kautz, J., Anandkumar, A.:
  Convolutional tensor-train lstm for spatio-temporal learning. In: Larochelle,
  H., Ranzato, M., Hadsell, R., Balcan, M.F., Lin, H. (eds.) Advances in Neural
  Information Processing Systems. vol.~33, pp. 13714--13726. Curran Associates,
  Inc. (2020)

\bibitem{wang_predrnn_nodate}
Wang, Y., Gao, Z., Long, M., Wang, J., Yu, P.S.: {P}red{RNN}++: Towards a
  resolution of the deep-in-time dilemma in spatiotemporal predictive learning.
  In: Dy, J., Krause, A. (eds.) Proceedings of the 35th International
  Conference on Machine Learning. Proceedings of Machine Learning Research,
  vol.~80, pp. 5123--5132. PMLR (10--15 Jul 2018)

\bibitem{xiao_spatiotemporal_2019}
Xiao, C., Chen, N., Hu, C., Wang, K., Xu, Z., Cai, Y., Xu, L., Chen, Z., Gong,
  J.: A spatiotemporal deep learning model for sea surface temperature field
  prediction using time-series satellite data. Environmental Modelling \&
  Software  \textbf{120},  104502 (Oct 2019)

\bibitem{zhang_gpu-acceleration_nodate}
Zhang, H., Si, S., Hsieh, C.J.: {GPU}-acceleration for {Large}-scale {Tree}
  {Boosting} p.~3

\end{thebibliography}
\end{document}